\documentclass[letterpaper]{article} 
\usepackage{aaai25}  
\usepackage{times}  
\usepackage{helvet}  
\usepackage{courier}  
\usepackage[hyphens]{url}  
\usepackage{graphicx} 
\usepackage{natbib}  
\usepackage{caption} 
\usepackage{algorithm}
\usepackage{algorithmic}

\usepackage{amsmath}
\usepackage{bm}
\usepackage{amssymb}
\usepackage{subcaption}
\usepackage{multirow}
\usepackage{booktabs}

\newtheorem{theorem}{Theorem}

\newtheorem{proposition}{Proposition}
\newtheorem{definition}{Definition}

\usepackage{newfloat}
\usepackage{listings}
\DeclareCaptionStyle{ruled}{labelfont=normalfont,labelsep=colon,strut=off} 
\floatstyle{ruled}
\newfloat{listing}{tb}{lst}{}
\floatname{listing}{Listing}
\nocopyright
\title{OU-CoViT: Copula-Enhanced Bi-Channel Multi-Task Vision Transformers \\ with Dual Adaptation for OU-UWF Images}
\author{
    Yang Li\textsuperscript{\rm 1}\thanks{Co-first authors.},
    Jianing Deng\textsuperscript{\rm 2}$^{*}$,
    Chong Zhong\textsuperscript{\rm 2}$^{*}$,
    Danjuan Yang\textsuperscript{\rm 3},
    Meiyan Li\textsuperscript{\rm 3},
    A.H. Welsh\textsuperscript{\rm 4},
    Aiyi Liu\textsuperscript{\rm 5},
    Xingtao Zhou\textsuperscript{\rm 3}$^{\dagger}$,
    Catherine C. Liu\textsuperscript{\rm 2}$^{\dagger}$,
    Bo Fu\textsuperscript{\rm 1}\thanks{Co-corresponding authors.}
}
\affiliations{
    \textsuperscript{\rm 1}School of Data Science,
  Fudan University\\
    \textsuperscript{\rm 2}Department of Applied Mathematics,
  The Hong Kong Polytechnic University\\
    \textsuperscript{\rm 3}Eye Institute and Department of Ophthalmology,
  Eye \& ENT Hospital, Fudan University\\
    \textsuperscript{\rm 4}College of Business and Economics,
  Australian National University\\
    \textsuperscript{\rm 5}Biostatistics and Bioinformatics Branch,
  Eunice Kennedy Shriver National Institute of Child Health and Human Development

    18110980006@fudan.edu.cn,
djnjoker@mail.ustc.edu.cn,
chzhong@polyu.edu.hk,
\{docdanjuanyang, limeiyan0406073\}@126.com,
Alan.Welsh@anu.edu.au,
liua@mail.nih.gov,
doctzhouxingtao@126.com,
macliu@polyu.edu.hk,
fu@fudan.edu.cn
}

\usepackage{bibentry}

\begin{document}

\maketitle

\begin{abstract}
Myopia screening using cutting-edge ultra-widefield (UWF) fundus imaging and joint modeling of multiple discrete and continuous clinical scores presents a promising new paradigm for multi-task problems in Ophthalmology.
The bi-channel framework that arises from the Ophthalmic phenomenon of ``interocular asymmetries'' of both eyes (OU) calls for new employment on the SOTA transformer-based models. 
However, the application of copula models for multiple mixed discrete-continuous labels on deep learning (DL) is challenging.
Moreover, the application of advanced large transformer-based models to small medical datasets is challenging due to overfitting and computational resource constraints. 
To resolve these challenges, we propose OU-CoViT: a novel Copula-Enhanced Bi-Channel Multi-Task Vision Transformers with Dual Adaptation for OU-UWF images, which can i) incorporate conditional correlation information across multiple discrete and continuous labels within a deep learning framework (by deriving the closed form of a novel Copula Loss); ii) take OU inputs subject to both high correlation and interocular asymmetries using a bi-channel model with dual adaptation; and iii) enable the adaptation of large vision transformer (ViT) models to small medical datasets. 
Solid experiments demonstrate that OU-CoViT significantly improves prediction performance compared to single-channel baseline models with empirical loss. 
Furthermore, the novel architecture of OU-CoViT allows generalizability and extensions of our dual adaptation and Copula Loss to various ViT variants and large DL models on small medical datasets. 
Our approach opens up new possibilities for joint modeling of heterogeneous multi-channel input and mixed discrete-continuous clinical scores in medical practices and has the potential to advance AI-assisted clinical decision-making in various medical domains beyond Ophthalmology.
\end{abstract}

%

\section{Introduction}
Myopia is a growing public health concern worldwide\cite{holden2016global}, and early detection and intervention through effective screening programs are crucial for preventing the progression of myopia and associated complications\cite{saw2005myopia}. 
Traditional screening methods rely on invasive, time-consuming evaluation and skilled professionals \cite{coan2023automatic}.
Recent advancements in non-invasive ultra-widefield (UWF) fundus imaging have opened up new possibilities for myopia screening which provides a comprehensive view of the retina with an expansive 200$^{\circ}$ field of view, enabling the detection of subtle changes associated with myopia.
The integration of UWF images with deep learning (DL) has the potential to revolutionize myopia screening by automating the process and improving its accuracy and efficiency, which can benefit telemedicine applications in rural areas lacking specialized ophthalmologists and medical resources \cite{ohsugi2017accuracy}.
However, current DL models for myopia screening faces several limitations:

Firstly, in DL applications on Ophthalmology, predicting discrete and continuous clinical scores such as binary high myopia (HM) status and Axial Length (AL) is of significant importance in Ophthalmology \cite{meng2011axial,zadnik2015prediction,tideman2016association}. 
However, most existing work only focused on the prediction of a single label and ignored their inherently high correlation. 
Recently, \cite{zhong2023cecnn} and \cite{li2024oucopula} demonstrated that joint modeling of correlated labels can improve the predictive capability of DL models by utilizing the conditional correlation information across labels. 
Nonetheless, the former can only model two labels and the latter can only deal with continuous labels. 
For multiple mixed discrete-continuous labels, a key challenge is to formulate a loss that is feasible for a DL training framework and characterizes the complicated conditional correlation structure of labels simultaneously. 
To the best of our knowledge, we are the first to resolve this challenge and successfully apply to the UWF image dataset.



Secondly, existing work rarely considered ``interocular asymmetries'' in Oculus Uterque (OU, both eyes) modeling. 
Interocular asymmetries, which refer to asymmetrical or unilateral features between left eyes (OS) and right eyes (OD) \cite{lu2022interocular}, imply that the fundus images of OU from the same patient may contain inconsistent information regarding the status of myopia. 
Studies proved that incorporating interocular asymmetries to the unilateral analysis could reduce statistical bias and provide additional information about retinal diseases \cite{sankaridurg2013correlation,henriquez2015intereye}.
Nonetheless, modeling OU in a DL model is difficult since the input OU images are strongly correlated yet exhibit heterogeneity due to interocular asymmetries.
Our solution is to develop a novel bi-channel model, which can simultaneously preserve the common features of OU and independently learn heterogeneous information contained in each eye.



Thirdly, due to the difficulty and expense of acquiring and annotating medical images, the limited data size hinders further applications of SOTA models of vision Transformer (ViT) and its variants on medical datasets \cite{li2020domain}.
If the data size is far much smaller than the model size, overfitting and computational resource constraints become serirous chanllenges. 
To overcome them, we adopt an idea from transfer learning and apply adaptation methods to pretrained models (e.g. from ImageNet) and fine-tune on our small UWF datasets.
This technique leverages the knowledge learned from the large dataset, reducing the risk of overfitting and the computational burden associated with training from scratch. 

By addressing the above limitations, we propose a novel framework \textbf{OU-CoViT}: Copula-Enhanced Bi-Channel Vision Transformers with Dual Adaptation for OU-UWF Images. 
Our model incorporates three key innovations:

\begin{enumerate}
    \item A computationally feasible Copula Loss for 4-dimensional mixed classification-regression tasks by deriving the closed form of the joint density, enabling to capture the conditional dependence structure among labels. 
    
    \item  A novel bi-channel architecture with dual adaptation and a shared backbone that simultaneously models heterogeneity (interocular asymmetries) and high correlation among multi-channel inputs.
    \item A highly efficient and easily accessible Ophthalmology AI practice of application of ViT using Low-Rank Adaptation (LoRA) in our bi-channel model that addresses the problem of using large transformer variants with small medical datasets.
\end{enumerate}

We evaluate the performance of OU-CoViT on our UWF fundus image dataset for myopia screening and demonstrate its superiority over single-eye-based baseline models. 
Furthermore, our dual adaptation architecture and Copula Loss function can be easily extended to other transformer variants and DL models.
Our approach not only opens up new possibilities for joint modelingcorrelations between discrete and continuous labels in multi-task learning but also has the potential to advance AI-assisted clinical decision-making in various medical domains beyond ophthalmology.

\section{Related work}
\paragraph{Deep learning with UWF fundus imaging}
Recently, DL techniques have been widely applied to UWF fundus imaging since it provides a comprehensive view of the retina which allows for better detection and monitoring of peripheral retinal diseases.
The many DL approaches that have been proposed to predict and detect retinal diseases from UWF fundus images have primarily focused on the diagnosis, classification, and segmentation of diseases \cite{li2021deep,zhang2021deepuwf,ju2021leveraging,engelmann2022detecting}.
In contrast, there has been limited research on the direct prediction of myopia and only a few myopia screening models have been developed with most of them mainly focused on binary classifications \cite{yang2020automatic,choi2021deep}.

\paragraph{Multi-task learning with correlation}
Multi-instance learning (MIL) and multi-task learning (MTL) are essential areas of machine learning that deal with problems with multiple labels. 
In MIL, the correlations between labels play a significant role in improving classification performance. However, most existing literature in MIL only considers the multiple classification problem \cite{song2018deep,chen2021learning,wu2023ctranscnn,lai2023single}, and many of them consider the spatial correlation across labels determined by small patches/instances in the image, rather than conditional correlation across labels.
MTL based on fundus images mostly focuses on disease classification \cite{sun2022multi,al2024fundus}. 
Only two ophthalmology works  considered the conditional correlation across labels \cite{zhong2023cecnn,li2024oucopula}, while their methods are only available on  traditional convolutional neural networks for some specific types of labels.


\paragraph{Adaptation methods}
The application of large transformers to medical datasets often faces two challenges: i) the limited size of available data hinders the training of large models from scratch; ii) the diverse range of diseases and use cases in medical scenarios makes it unaffordable to retrain the entire large model for each specific medical task. 
Adaptation methods offer a solution to both of these problems by enabling the efficient fine-tuning of pre-trained models for various tasks and domains, such as TAIL \cite{liu2023tail}, G-adapter \cite{gui2024g}, Bitfit \cite{zaken2021bitfit}, AdaptFormer \cite{chen2022adaptformer} and LoRA \cite{hu2021lora}.
In medical applications, ``MeLo''\cite{zhu2023melo} which applied LoRA for medical image diagnosis across various clinical tasks, achieved satisfactory results with significant reductions in storage space and computational requirements. 
However, current research lacks a specific architecture that can simultaneously handle the transfer of large models to small datasets and address the heterogeneity of multi-channel data.

\section{Methods}
\begin{figure*}[!htb]
    \centering\includegraphics[width=.9\textwidth]{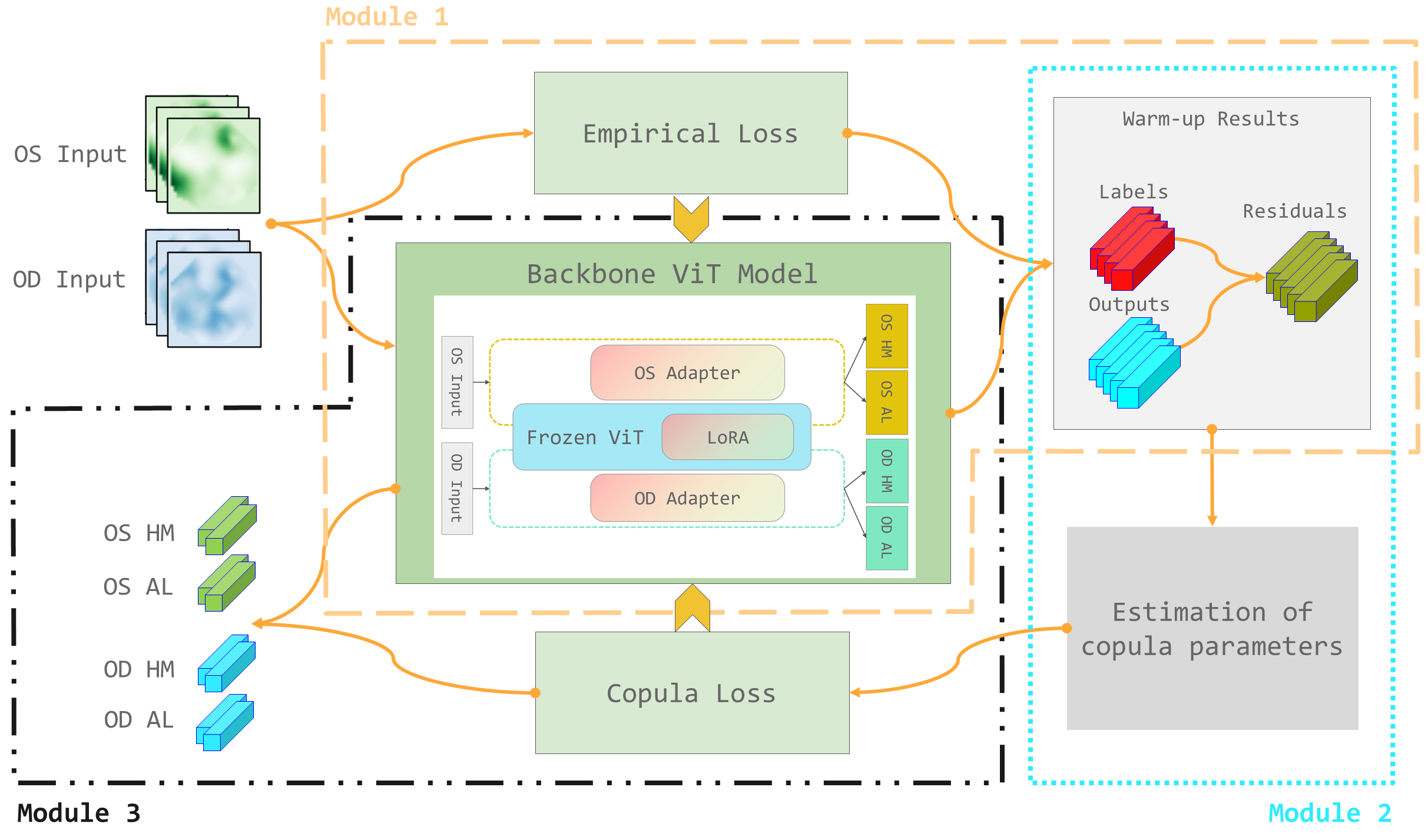}
    \caption{Architecture of proposed OU-CoViT. 
    } \label{fig:diagram_OUCoViT}
\end{figure*}

The proposed OU-CoViT has two essence components, the Copula Loss and dual adaptation. 
The OU-CoViT is implemented through three modules, as shown in  Fig.~\ref{fig:diagram_OUCoViT}. 
We introduce the two essence components in sections \ref{subsec:copula_modeling} and \ref{subsec:dual_adaptation} respectively, and describe the the whole architecture of OU-CoViT in details in section \ref{subsec:Endtoend_OUCoViT}. 

\subsection{Copula modeling}\label{subsec:copula_modeling}
\subsubsection{Gaussian copula for multi-label regression-classification tasks}
Let $\bm{y} = (y_{1}, y_{2}, y_{3}, y_{4})$ be the 4-dimensional vector of mixed discrete-continuous responses, where $y_{1}, y_{2} \in \mathbb{R}_+$ correspond to OS AL and OD AL, and $y_{3}, y_{4} \in \{0, 1\}$ correspond to OS high myopia status and OD high myopia status (1: high myopia; 0, otherwise), respectively. 
Let $\mathcal{X} \in \mathbb{R}^{224 \times 224 \times 3}$ be the UWF image predictors. 
We have the following generative model
\begin{align}
\label{MarginalModel}
\begin{split}
y_{m}|\mathcal{X} &\sim N(g_{m}(\mathcal{X}),\sigma_{m}^2), \\
~y_{m+2} &\sim \text{Bernoulli}[\text{Sigmoid}\{g_{m+2}(\mathcal{X})\}], ~ m=1, 2, 
\end{split}
\end{align}
where $g_{m}$ denotes the unknown regression function associating the predictor $\mathcal{X}$ with label $y_{m}$.

We adopt the copula model \cite{sklar1959fonctions} to 
capture the conditional correlation $corr(y_{i1}, \ldots, y_{i4} | \mathcal{X}_i)$. 
In general, a $p$-dimensional copula is a parametric function $C$ on $[0, 1]^p$ such that the joint distribution 
$$
F(y_1, \ldots, y_p) = C\{F_1(y_1), \ldots, F_p(y_p)\}, 
$$
where $F_j$ denotes the $j$th marginal cumulative distribution function (CDF) of $y_j$ for $j=1, \ldots, p$. 
Specifically, in this paper, we consider the Gaussian copula model 
\begin{align}
\label{GaussianCopula}
    C(\bm{y}|\Gamma) = \Phi_p(\Phi^{-1}\{F_1(y_1)\}, \ldots, \Phi^{-1}\{F_p(y_p)\}|\bm{0}, \Gamma), 
\end{align}
where $\Phi_p$ is the CDF of a $p$-dimensional Gaussian distribution equipped with a correlation matrix $\Gamma \equiv (\gamma_{tj})_{p\times p}$, and $\Phi^{-1}$ is the inverse CDF of the standard normal distribution.

\subsubsection{4-dimensional Copula Loss}
Commonly used empirical losses for multiple regression-classification tasks are MSE and cross-entropy, which are equivalent to the negative log densities of the marginal model \eqref{MarginalModel}. 
However, these losses ignore the the conditional correlation $corr(y_{i1}, \ldots, y_{i4} | \mathcal{X}_i)$.
Consequently, a loss based on the joint density of the Gaussian copula model \eqref{GaussianCopula} is needed. 
Under model \eqref{GaussianCopula}, the general form of the joint density is given by \cite{song2009joint}
\begin{align}
\label{PeterSong: jointdensity}
\begin{split}
&f(\bm{y})=f_1(y_{1})f_2(y_{2})\\
&\sum_{j_{3}=1}^{2} \sum_{j_4=1}^{2} (-1)^{j_{3}+ j_{4}}C_1^{2} (F_1(y_1),F_{2}(y_{2}),u_{3,j_3},u_{4,j_4}|\Gamma),
\end{split}
\end{align}
where $u_{j,1} = F_j(y_j-)$, $u_{j,2} = F_j(y_j)$, $F_j(y_j-)$ is the left-hand limit of $F_j$ at $y_j$, and 
\begin{align}
\begin{split}
\label{PeterSong: copuladerivative}
    &C_1^{2}(u_1,u_2,u_3,u_4|\Gamma)=\frac{\partial^{2}}{\partial u_1 \partial u_2}C(u_1,u_2,u_3,u_4|\Gamma)\\
 &\propto\int_{-\infty }^{\Phi^{-1}(u_{3})}  \int_{-\infty }^{\Phi^{-1}(u_4)}\\
 &\exp\left \{-\frac{1}{2}(\bm{q}^T,\bm{x}^T)\Gamma^{-1}(\bm{q}^T,\bm{x}^T)^T 
 +\frac{1}{2}\bm{q}^T\bm{q}    \right \} \mathrm{d}x_4\mathrm{d}x_3, 
\end{split}
\end{align}
where $\bm{q} = (\Phi^{-1}(u_1),\Phi^{-1}(u_2))^T$ and $\bm{x} = (x_3,x_4)^T$.
Nonetheless, the integral expression \eqref{PeterSong: copuladerivative} is non-trivial, hindering DL training with this loss. 
Although \cite{onken2016mixed} provides convenient computation of the likelihood under the Gaussian copula, their method introduces infinity during backward propagation, so cannot be used to train a DL model. 
To resolve this problem, the following theorem provides the closed form of the joint density \eqref{PeterSong: jointdensity}.  
The proof is deferred to Supplementary Materials
. 



\begin{theorem}
\label{theorem: jointdensity}
Suppose that marginally $y_j \sim  N(\mu_j,\sigma_j^2)$, for $j=1,2$, $y_{3} \sim \text{Bernoulli}(p_3)$, and $y_{4} \sim \text{Bernoulli}(p_4)$. 
Let 
$$
\Gamma = \begin{pmatrix} 1 & \gamma_{12} & \gamma_{13} & \gamma_{14}\\\gamma_{12} & 1 & \gamma_{23} & \gamma_{24}\\\gamma_{13} & \gamma_{23} & 1 & \gamma_{34}\\\gamma_{14} & \gamma_{24} & \gamma_{34} & 1\\\end{pmatrix}
$$
be the correlation matrix in the Gaussian copula \eqref{GaussianCopula}. 
Then the closed form of the log joint density \eqref{PeterSong: jointdensity} is

\begin{align}
\begin{split}
\label{closeform: jointdensity}
&f(y_1,y_2, y_3, y_4) =
-\frac{1}{2}\bm{q}^T\Gamma_{11}^{-1}\bm{q} + \\
&log(\sum_{j_{3}=1}^{2}\sum_{j_4=1}^{2} (-1)^{j_{3}+j_{4}}C^{*} (u_{{3},j_3},u_{{4},j_4})) + C, 
\end{split}
\end{align}
where $\bm{q} = (\frac{y_1-\mu_1}{\sigma_1},\frac{y_2-\mu_2}{\sigma_2})^T$, $C$ is a constant and the values of function $C^*$ are listed in the table 1 of Lemma 3 in Supplementary Materials.

\end{theorem}

Finally, the Copula Loss is minus the log-likelihood, so based on Theorem \ref{theorem: jointdensity}, we obtain
\begin{equation}
\label{copulaloss}
\begin{split}
\mathcal{L}(g_{1},g_{2},p_{3},p_{4}) = - \sum_{i=1}^n(-\frac{1}{2}\bm{q}_i^T\Gamma_{11}^{-1}\bm{q}_i + \\
log(\sum_{j_{3}=1}^{2}\sum_{j_4=1}^{2} (-1)^{j_{3}+j_{4}}C^{*} (u_{i3,j_3},u_{i4,j_4}))).
\end{split}
\end{equation}

\subsubsection{Estimation of copula parameters}\label{subsubsec:Parameter_estimation}
The Copula Loss \eqref{copulaloss} is parameterized by $(\sigma_1, \sigma_2, \Gamma)$. 
It is difficult to learn these parameters by direct optimization. 
Nonetheless, under the Gaussian copula, each of them has a clear statistical interpretation. 
Hence, we provide details of their estimation here. 

We first introduce the definition of Gaussian scores for both continuous and discrete responses. 
\begin{definition}
Let $X \in \{0, 1\}$ be a binary random variable. 
Suppose there exists a latent standard Gaussian variable $Z$ 
such that $Pr\{X=1\} = p_0 = Pr\{Z \le z\}=\Phi(z)$. 
We call $z$ the Gaussian score of $X$. 

Let $Y$ be a continuous random variable with CDF $F(y)$. 
We call $q(y) = \Phi^{-1}(F(y))$ the Gaussian score of $Y$. 
\end{definition}

Theorem \ref{theorem: jointdensity} yields the following proposition. 
The proof is deferred to Supplementary Materials.
\begin{proposition}
\label{prop: Gaussianscores}
Suppose the  random vector $\bm{y} = (y_1, y_2, y_3, y_4)$ has log joint density in the form of \eqref{closeform: jointdensity} with correlation matrix $\Gamma$. 
Let  $\bm{q} = \{q(y_1), q(y_2)\} = \{q_1, q_2\}$ and $\bm{z} = \{z_3, z_4\} $ be the  Gaussian scores of $(y_1, y_2)$ and $(y_3, y_4)$, respectively.  
We have 
$$
(\bm{q}, \bm{z}) \sim N_4(\bm{0}, \Gamma). 
$$
\end{proposition}

Proposition \ref{prop: Gaussianscores} tells that under the Gaussian copula, the correlation between Gaussian scores is the same as the correlation structure of the copula. 
Therefore, it is natural to estimate  all the elements in the correlation matrix $\Gamma$ using the Pearson correlation of the Gaussian scores. 
That is, 
\begin{align*}
    &\hat{\gamma}_{12} = corr(z_1,z_2), \\
&\hat{\gamma}_{34} = corr(\Phi^{-1}(\text{Sigmoid}\{g_{3}(\mathcal{X})),\Phi^{-1}(\text{Sigmoid}\{g_{4}(\mathcal{X}))), \\
&\hat{\gamma}_{t,(j+2)} = corr(z_t,\Phi^{-1}(\text{Sigmoid}\{g_{j+2}(\mathcal{X}))),~t,j = 1,2
\end{align*}
where $z_t = \frac{y_{t}-g_{t}(\mathcal{X})}{\sigma_t}$ is the standardized residual.

By definition, $\sigma_1^2$ and $\sigma_2^2$ correspond to the marginal variances of $y_1$ and $y_2$, respectively, so it is natural to estimate $(\sigma_1, \sigma_2)$ by the sample standard deviation of their marginal residuals. 


\subsection{Dual adaptation}\label{subsec:dual_adaptation}
\begin{figure*}[tb]
\centering
\begin{subfigure}{0.22\textwidth}
    \includegraphics[width=\textwidth]{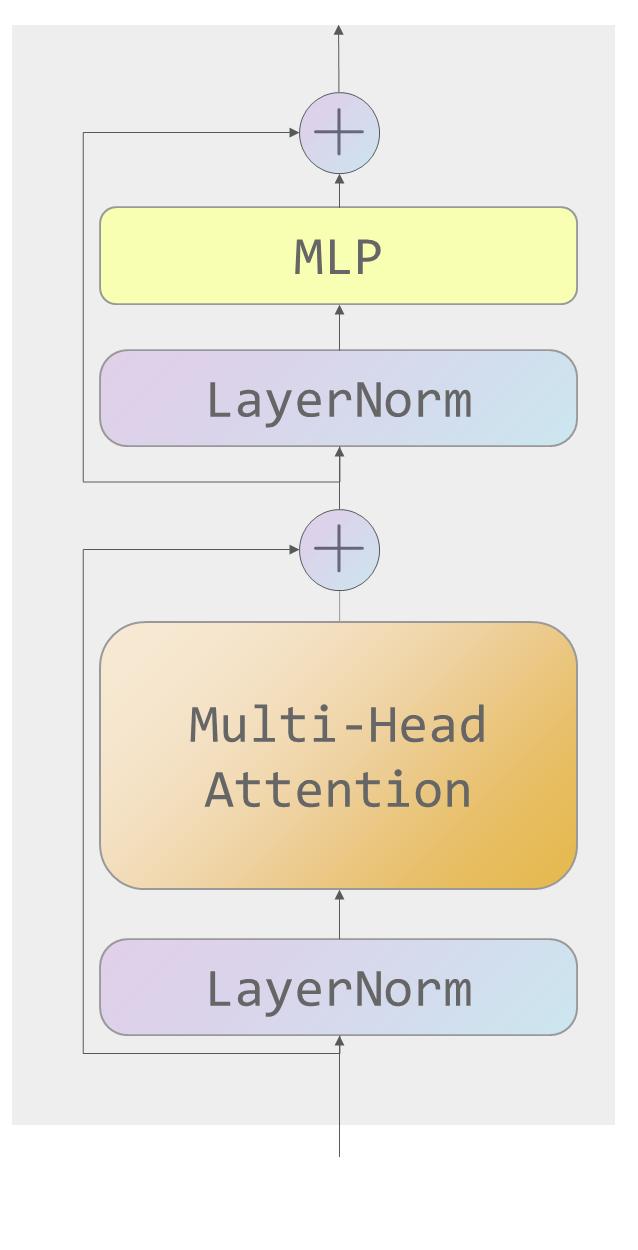}
    \caption{} 
    \label{fig:diagram_vitblock}
\end{subfigure}
\begin{subfigure}{0.7\textwidth}
    \includegraphics[width=\columnwidth]{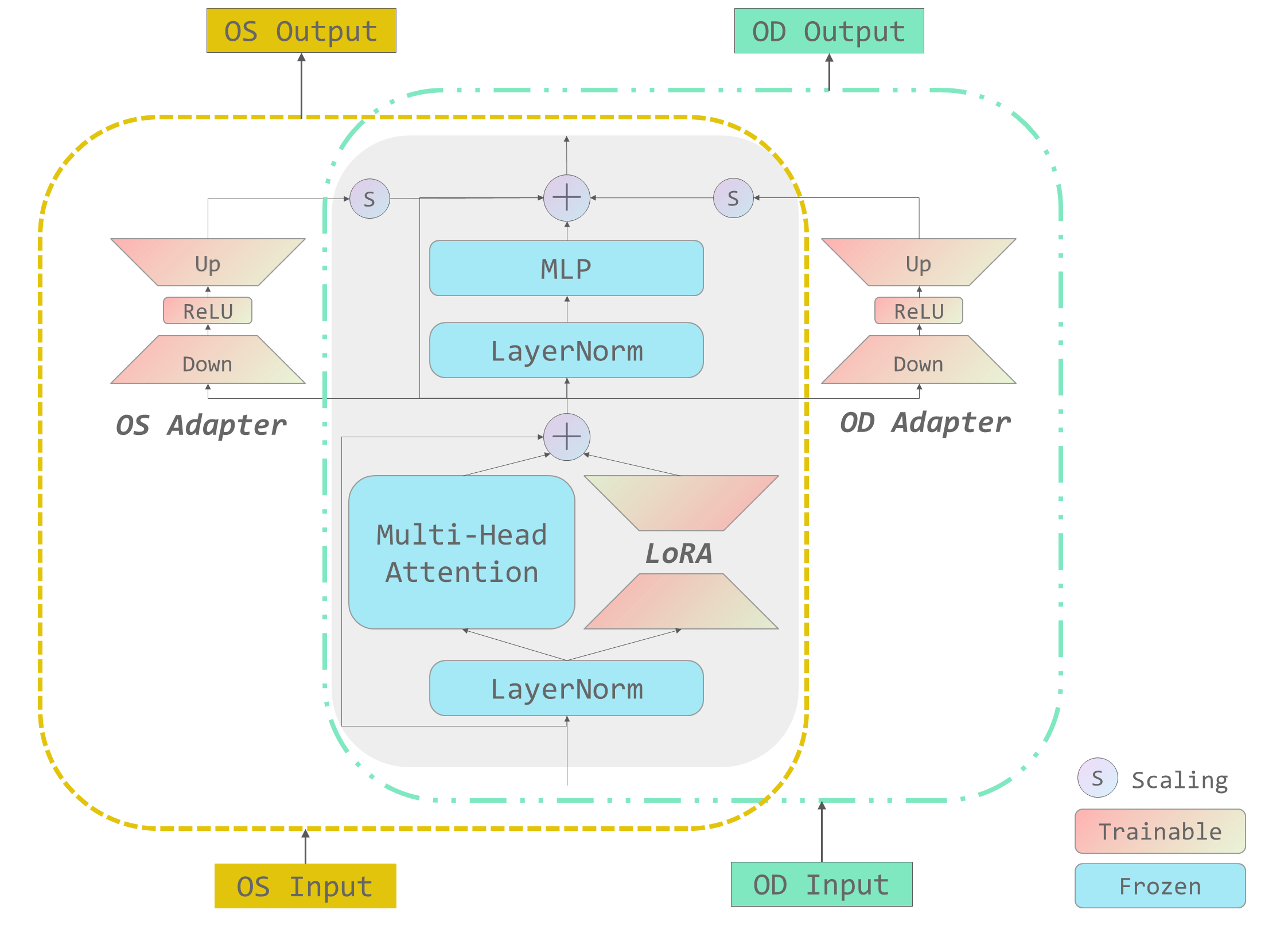}
    \caption{} 
    \label{fig:dualAdaptArchitect}
\end{subfigure}
\caption{(a) Original transformer block in classic single-channel ViT; (b) Detailed architecture of proposed dual adaptation in one transformer block with bi-channel modeling.}
\end{figure*}

We propose a novel dual adaptation bi-channel structure to address the issue of interocular asymmetries in OU images and the challenges of the small size of the UWF dataset.
Inspired by prior literature, we adopt two adaptation methods: the adapter from AdaptFormer \cite{chen2022adaptformer} and LoRA \cite{hu2021lora}. 
The adapters capture the heterogeneous information arising from the interocular asymmetries, while LoRA enables transfer learning on the large-scale ViT. 
Fig.~\ref{fig:diagram_vitblock} and Fig.~\ref{fig:dualAdaptArchitect} illustrate the detailed architecture of our dual adaptation method and the comparison between our bi-channel architecture and conventional single-channel transformer blocks.

\subsubsection{Adapter modules for interocular asymmetries}
UWF images of both eyes were jointly modeled firstly in Ophthalmology by \cite{li2024oucopula}, where they may be the first to employ adapters in a bi-channel architecture in a DL model.  
The adapter structure was first introduced in \cite{houlsby2019parameter}, and recently this classic yet simple design has been widely adopted in many adaptation methods.
We adopt adapters for a novel bi-channel architecture for ViT to accommodate the high correlation and interocular asymmetries between OU inputs.
Unlike traditional approaches that assign separate neural networks to OS and OD, our framework aims to share most parameters in the backbone to preserve most of the common features between OU, while allowing small but distinct adapters to learn the heterogeneous information for each eye.

We constructed adapter modules for both the left and right eyes (OS Adapter and OD Adapter, respectively) and let them share the backbone model. 
Fig.~\ref{fig:dualAdaptArchitect} shows the detailed architecture of OS and OD Adapter.
During training, adapters are updated simultaneously with the backbone model. 
We inserted the OS and OD Adapters into the MLP module of each transformer block in a manner similar to AdaptFormer. This approach offers several advantages: 1) The multi-head attention modules of different transformer variants typically have different structures, but their MLP modules are similar. This means our adapter modules can be inserted into transformer variants beyond ViT;
2) Asymmetrical information between the eyes usually reflects differences in disease severity or progression rather than entirely different manifestations of the same condition. Thus, the feature extraction step via multi-head attention can be shared, while the feature transformation should differ slightly between OU;
3) This is a straightforward and classical structure, making it easy for medical researchers to implement and facilitating further research.

\subsubsection{LoRA for transfer learning}
To apply large models to small datasets, we adopted LoRA for our bi-channel ViT. 
The core concept of LoRA involves freezing the pre-trained model weights and incorporating trainable low-rank decomposition matrices into each layer of the Transformer architecture, which can substantially reduce the number of trainable parameters. 

In our approach, we employ LoRA to effectively adapt the ViT base-sized model pretrained on ImageNet-21k \cite{wu2020visual,deng2009imagenet} to our bi-channel model.
As shown in Fig.~\ref{fig:dualAdaptArchitect}, we applied LoRA to the multi-head attention (MHA) module, using blue and gradient pink colors to distinguish between frozen and trainable parameters. The advantage of applying LoRA to the MHA module lies in the maturity of existing software support, enabling medical researchers to use LoRA with various transformer variants. Additionally, we selected the base-sized ViT  because it is easily accessible to medical researchers and can be loaded on one GTX4090 card.

\subsection{End-to-end OU-CoViT}\label{subsec:Endtoend_OUCoViT}
OU-CoVit consists of three modules: i) a warm-up module that trains the backbone bi-channel ViT under empirical losses (cross entropy and MSE loss for our experiments); ii) a copula estimation module that estimates the parameters in the Copula Loss based on the outputs from the warm-up module; iii) the OU-CoViT module that trains the backbone bi-channel ViT using the derived Copula Loss. We summarize each step of the OU-CoViT in Algorithm 1 in Supplementary Materials.

For labels $\bm{y} = (y_1, y_2, y_3, y_4)$ marginally generated from model \eqref{MarginalModel}, Module 1 provides maximum likelihood estimators $(\hat{y}_1, \hat{y_2})$ and $(\hat{p}_3, \hat{p}_4)$. 
The resulting residuals and Gaussian scores are then imported to Module 2 to estimate the copula parameters. 
Finally, in Module 3, we train the backbone model under Copula Loss with copula parameters estimated in Module 2. 
During the training process in Module 3, the OS input passes through the transformer blocks with LoRA and the OS Adapters, while the OD input goes through the transformer blocks with LoRA and the OD Adapters. Each training iteration involves inputting images from both eyes of the same individual. Subsequently, the outputs from both eyes are used together in the computation of the Copula Loss, followed by backward propagation to complete a single training iteration. 
Note that the Copula Loss does not rely on any specific architecture of the backbone model.

\section{Experiments}
In this section we evaluate OU-CoViT by investigating whether the proposed framework can improve the baseline model by extracting conditional correlation information for the outcomes and addressing the issue of interocular asymmetries. 
We also conducted a series of ablation experiments to validate the effectiveness and to understand the impact of the various components of OU-CoViT .

\paragraph{Dataset} 
The data collection process involved capturing 5,228 UWF fundus images from the eyes of 2614 patients using the Optomap Daytona scanning laser ophthalmoscope (Daytona, Optos, UK). 
All enrolled patients sought refractive surgery treatment and were exclusively myopia patients. 
The data collection period extended from Dec. 2014 to Jun. 2020, and was conducted at The Eye \& ENT Hospital of Fudan University. 
The UWF fundus images obtained during the study were exported in JPEG format and compressed to a resolution of 224 x 224 pixels to facilitate subsequent analysis.

\paragraph{Experiment setup and evaluation metrics}
The dataset of 5,228 fundus images was partitioned into the training data set, the validation set, and the testing set, with a ratio of 6:2:2. 
To mitigate bias in model evaluation and obtain more reliable estimates of the results, 5-fold cross-validation (CV) was employed. 
Since the rank $r$ in LoRA significantly affects the number of trainable parameters and the model's performance, we tested the results for $r=4, 8, 16$. We limited $r$ to a maximum of 16 because larger values resulted in an excessive number of trainable parameters, leading to severe overfitting of the ViT model on our UWF dataset.
We calculate MSE for continuous labels, cross entropy (CE) and Area Under the Curve (AUC) for the discrete labels.  
Specifically, the following results for each LoRA rank $r=4, 8, 16$ were compared: 
1) Average AUC \& cross entropy of HM, MSE of AL for each eye (AUC HM OS/OD, Cross Entropy HM OS/OD, \& MSE AL OS/OD);
2) Average AUC \& cross entropy of HM, MSE of AL for both eyes (AUC HM OU, Cross Entropy HM OU, \& MSE AL OU);


We compared the performance between ViT with LoRA (baseline model), ViT + LoRA + Adapters (dual adaptation only), ViT + LoRA + Copula Loss (copula only), and ViT + dual adaptation + Copula Loss (OU-CoViT), aiming to show that each of the innovative modules/components we proposed is effective and makes an important contribution.
For the baseline ViT with LoRA, since there is no adapter module, the bi-channel model is then simplified to a single-channel ViT, which can be treated as a single-eyed model. 

\paragraph{Implementation details}
We utilize a pretrained ViT base-sized model pretrained on ImageNet-21k \cite{wu2020visual,deng2009imagenet} to train our OU-CoViT.
We trained the backbone model in Module 1 with 20 epochs, and for OU-CoViT, we trained it with 15 epochs.
The batch size of 32 is used due to memory limitation.
The model is optimized with Adam \cite{kingma2014adam} with an initial learning rate 1e-4 and reduced to 1e-5 after 10 epochs.
From a medical perspective, the heterogeneity caused by interocular asymmetry is relatively small, though not negligible. Therefore, the bottleneck middle dimension $\hat{d}$ of the adapter is set to a small value of $\hat{d}=1$. For the selection of the inserted layers and the scaling factor of the adapter module, we followed the recommendations from the AdaptFormer \cite{chen2022adaptformer}, where adapters are inserted into every transformer block, and the scaling factor is set to 0.1.
All experiments are done with a single RTX 4090 24GB card.

\subsection{Results on UWF dataset}
\begin{figure*}[tb]
    \centering\includegraphics[width=1.0\textwidth]{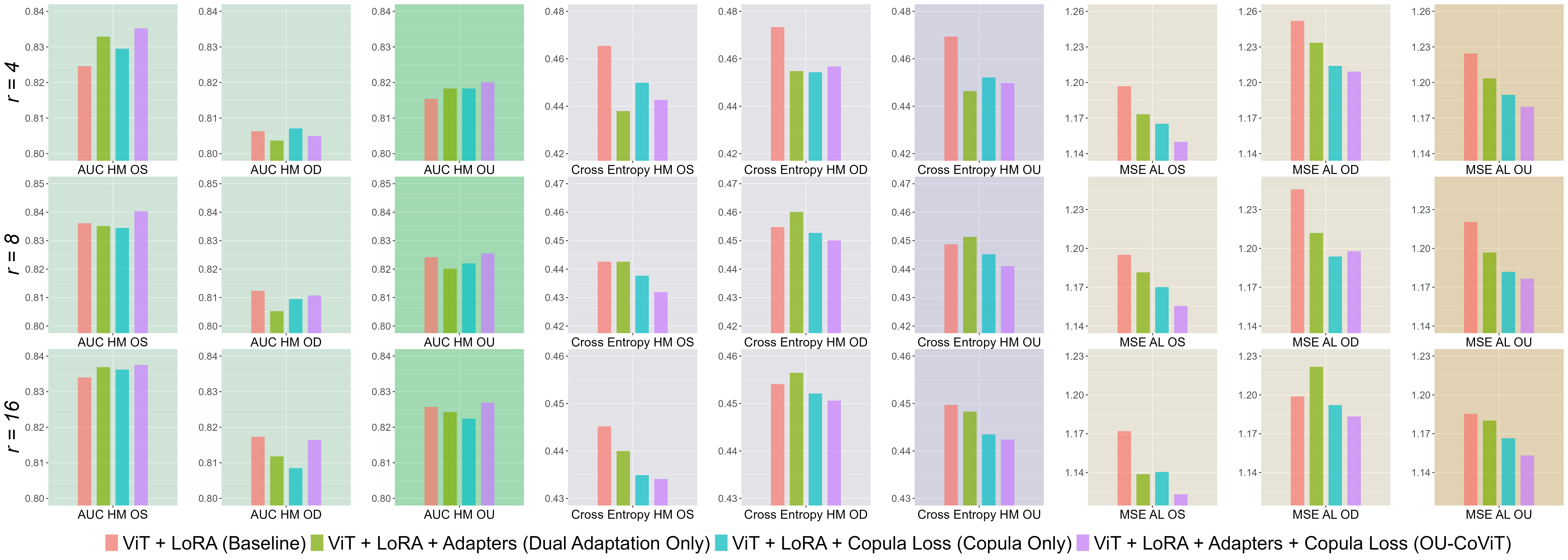}
    \caption{Prediction performances of the UWF dataset under different LoRA ranks ($r=4,8,\& 16$). 
    } \label{fig:main_results}
\end{figure*}

In this section, we present the results of our experiments, which evaluate the performance of our proposed OU-CoViT framework across various metrics and configurations. 
Fig.~\ref{fig:main_results} provides a comprehensive comparison between ViT with LoRA (baseline model), ViT + LoRA + Adapters (dual adaptation only), ViT + LoRA + Copula Loss (copula only), and ViT + dual adaptation + Copula Loss (OU-CoViT) under different LoRA rank parameters $r=4, 8, 16$.
Notably, higher AUC results are better, while lower cross entropy and MSE results are preferable.
The results consistently demonstrate that OU-CoViT generally improves performance of sole baseline models and backbones with dual adaptation or Copula Loss solely across all three ranks.
It is also worth noting that as the rank $r$ increases, the overall performance improves slightly, as a higher rank in the LoRA model leads to more trainable parameters and, consequently, stronger predictive capability of the model.

For the regression results, dual adaptation, the Copula Loss, as well as OU-CoViT, all significantly improve the MSE of AL. 
Among these, OU-CoViT provides the strongest enhancement.
For the classification results, OU-CoViT shows significant improvement in cross entropy, with the greatest gain observed at $r=4$. Although the improvement in AUC is less pronounced, it is still evident. 
Both dual adaptation and Copula Loss individually also enhance cross entropy; their improvement on AUC is most notable at $r=4$.

Therefore, we can conclude that both dual adaptation and Copula Loss, whether used separately or combined, enhance the predictive capability of the baseline model. 
The greatest improvement occurs when both methods are integrated in OU-CoViT.
It is worth noting that for myopia screening prediction, the average AL MSE of OU-CoViT surpasses previous results \cite{li2024oucopula} based on the same dataset using ResNet as the backbone (1.153 vs. 1.719). This indicates that the performance of transferring pre-trained large ViT-based model could be significantly better than that of traditional CNN-based models.

\subsection{Ablation study}

To demonstrate the generalizability of Copula Loss, proving that it can be applied to various DL models, we tested its effectiveness additionally on the ViT large-sized model \cite{wu2020visual,deng2009imagenet} and the simplified ResNet from \cite{li2024oucopula}. 
We aimed to see if Copula Loss could improve the predictive capability of these backbone models. 
The results in Table~\ref{table:ablation_copula} show that MSE is substantially improved with Copula Loss and the inclusion of Copula Loss slightly improves classification tasks.

To demonstrate the rationale behind the placement of the adapter module within the transformer block, we also tested several other common adapter positions, including adapting after the embedding layer, before the Feed-Forward Network (FFN) layer, and before the fully-connected (FC) layer across different LoRA ranks.
Table~\ref{table:ablation_adaptation} shows that although some adapter positions achieve better results, no configurations outperforms our method on both classification and regression tasks across all ranks $r$. 


\begin{table}[tb]
 
    \caption{{Generalizability of Copula Loss. ``W/ Copula'' means the corresponding backbone model is trained with Copula Loss.}}
    \label{table:ablation_copula}
     \centering
    \setlength{\tabcolsep}{1mm}{
    \begin{tabular}{cccccc}
    \toprule
    & \multicolumn{2}{c}{\textbf{Cross Entropy}} & & \multicolumn{2}{c} {\textbf{MSE}} \\
    \cmidrule{2-3}
    \cmidrule{5-6}
    & HM OS & HM OD  & & AL OS & AL OD  \\ 
    
    \midrule
        ViT Large & \textbf{0.4438} & 0.4590  & & 1.1846 & 1.2540  \\ 
        W/ Copula
        & 0.4439 & \textbf{{0.4587}}  & & \textbf{{1.1475}} & \textbf{{1.2249}} \\ 
        \midrule
        ResNet & 0.5962 & 0.5732 & & 3.0859 & 2.9807 \\ 
        W/ Copula & \textbf{{0.5959}} & \textbf{{0.5710}} & & \textbf{{2.6362}} & \textbf{{2.6741}}  \\ 
    \bottomrule
    
    \end{tabular}}
    \end{table}

\begin{table}[tb]
    \caption{{Effects of the position of adapters. CE is calculated based on HM of OU; MSE is calculated based on AL of OU.}}
    \label{table:ablation_adaptation}
     \centering
    \setlength{\tabcolsep}{1mm}
    \fontsize{9pt}{9pt}\selectfont
    
\begin{tabular}{ccccccccc}
    \toprule
   & \multicolumn{2}{c}{{\textit{r} = 4}} & & \multicolumn{2}{c} {\textit{r} = 8} & & \multicolumn{2}{c} {\textit{r} = 16} \\
    \cmidrule{2-3}
    \cmidrule{5-6}
    \cmidrule{8-9}
    & CE & MSE && CE & MSE & &CE & MSE \\ 
    \midrule
        \textbf{
        At FFN (ours)} & \textbf{\textit{0.451}} & \textbf{\textit{1.203}} & & \textbf{\textit{0.451}} & \textbf{\textit{1.197}} & & \textbf{\textit{0.448}} & \textit{\textbf{1.180}} \\ 
        After Embedding
        & 0.469 & 1.215 & & 0.451& 1.216 & & 0.449& 1.205 \\ 
        Before FFN  & \textbf{0.446} & \textbf{1.201} & & \textbf{0.444} & 1.208 & & 0.454& 1.202\\ 
        Before  FC   & 0.452 & 1.224 & & 0.462 & 1.230 & & \textbf{0.446} & 1.199\\
        
    \bottomrule
    
    \end{tabular}
    \end{table}

\section{Discussion}
We succeed in jointly predicting 4 important clinical scores in Ophthalmology using OU UWF fundus images simultaneously. 
Our success demonstrates two strengths of the proposed OU-CoViT: i) enhancing the predictive capabilities of baseline models for multiple mixed classification and regression tasks, and ii) dealing with bi-channel imaging inputs with high correlation and inherent heterogeneity under a ViT model.  
The integration of dual adaptation and Copula Loss proves to be a robust approach, with flexible generalizability to other transformer variants and DL models.

To the best of our knowledge, our work is the first to integrate the conditional correlation information across multiple discrete and continuous labels, which is suitable for multi-task learning involving mixed discrete-continuous tasks.
Our Copula Loss also opens up new avenues for research into more diverse correlation modeling in DL.

The improvements brought by dual adaptation indicate that the bi-channel architecture based on OU modeling has stronger predictive capabilities than the baseline model based on single-eye input.
Both of our adaptation methods can be easily inserted into various transformer variants, providing significant convenience for medical researchers. 
Additionally, although we only built a bi-channel model in this study, the adapter structure indicates that our framework can be extended to multi-channel learning problems.
This presents a novel strategy for handling heterogeneous multi-channel inputs.

Compared to results on ResNet, the results using ViT with transfer learning are apparently better, demonstrating the great potential of the LoRA method for addressing the challenge of applying large models to small medical datasets. 
Our framework is not limited to ViT and can be applied to different backbones according to various medical scenarios.
We demonstrate a new Ophthalmology AI practice of transfer learning, which provides researchers with new opportunities.  



\paragraph{Limitations}
The limitations of our study include the following: the closed-form Copula Loss we derived is case-specific and currently applicable to the UWF myopia screening model. 
However, this opens up new directions and opportunities for application and closed-form derivation in other fields. 
The definition of $\bm{q}$ in \eqref{closeform: jointdensity} indicates that the inverse of the marginal variance of AL can be viewed as the weight on the loss of regression tasks. 
Smaller variance of AL leads to a larger weight, yielding that OU-CoViT tends to optimize regression loss more than classification loss. 
A feasible future direction would be to improve this phenomenon to enable the model to optimize the loss of all labels in a more balanced way.

\section{Conclusion}
Our OU-CoViT framework, with its dual adaptation and Copula Loss, fills a gap in AI application to Ophthalmology that simultaneously deploys UWF fundus images of both eyes, and accordingly constructs a novel framework using the large ViT model on small datasets for multiple classification and regression tasks. 
Comprehensive experiments  and ablation studies validate the robustness and generalizability of our approach, paving the way for its application in diverse and challenging medical scenarios.

\bibliography{aaai25}

\end{document}